\documentclass[letterpaper, 10 pt, conference]{ieeeconf}  

\IEEEoverridecommandlockouts                              

\usepackage{amsmath} 
\usepackage{amssymb}  
\usepackage{subcaption}
\usepackage{graphicx}
\usepackage{algorithm}
\usepackage{algpseudocode}

\newtheorem{theorem}{Theorem}

\title{\LARGE \bf
Resampling-free Particle Filters in High-dimensions
}

\author{Akhilan Boopathy$^{1}$, Aneesh Muppidi$^{2}$, Peggy Yang$^{1}$, Abhiram Iyer$^{1}$, William Yue$^{1}$, Ila Fiete$^{1}$
\thanks{$^{1}$MIT, $^{2}$Harvard}%
}

\begin{document}

\maketitle
\thispagestyle{empty}
\pagestyle{empty}

\begin{abstract}

State estimation is crucial for the performance and safety of numerous robotic applications. Among the suite of estimation techniques, particle filters have been identified as a powerful solution due to their non-parametric nature. Yet, in high-dimensional state spaces, these filters face challenges such as 'particle deprivation' which hinders accurate representation of the true posterior distribution. This paper introduces a novel resampling-free particle filter designed to mitigate particle deprivation by forgoing the traditional resampling step. This ensures a broader and more diverse particle set, especially vital in high-dimensional scenarios. Theoretically, our proposed filter is shown to offer a near-accurate representation of the desired posterior distribution in high-dimensional contexts. Empirically, the effectiveness of our approach is underscored through a high-dimensional synthetic state estimation task and a 6D pose estimation derived from videos. We posit that as robotic systems evolve with greater degrees of freedom, particle filters tailored for high-dimensional state spaces will be indispensable.
\end{abstract}


\section{Introduction}

State estimation remains at the heart of many robotic applications, from autonomous navigation to manipulation tasks. Precise and timely estimates of the system's state are critical to ensure the safe and effective operation of robotic systems. Among state estimation techniques, particle filters have emerged as a particularly potent tool~\cite{thrun2005probabilistic}. Their non-parametric nature allows them to flexibly model a wide variety of distributions, making them ideal for capturing the intricacies of complex robotic systems.

However, using particle filters in high-dimensional state spaces has specific challenges. One issue is the so-called 'particle deprivation' or 'sample impoverishment'~\cite{arulampalam2002tutorial}. This refers to the scenario where large regions of the state space lose all representation during particle resampling, making it challenging to recover density in these regions later if necessary. In high-dimensional spaces, particle deprivation becomes increasingly problematic, often resulting in a particle filter that does not sufficiently represent the true posterior distribution.

In light of these challenges, this paper presents three primary contributions:
\begin{enumerate}
    \item We introduce a novel resampling-free particle filter that mitigates the issue of particle deprivation. By avoiding the conventional resampling step, our approach ensures a more diverse set of particles, enhancing representation in high-dimensional spaces.
    \item On the theoretical front, we rigorously show that our proposed particle filter provides a close approximation to the desired posterior distribution, even when particles lie in high-dimensional state spaces.
    \item We empirically demonstrate the efficacy of our method on a high-dimensional synthetic state estimation task and 6D pose estimation from videos.
\end{enumerate}

This paper aims to bridge the gap between particle filter theory and its applicability in high-dimensional robotic systems. By presenting a resampling-free approach, we hope to pave the way for more robust state estimation methods in complex robotic applications.

\section{Related Work}

Particle filters have gained immense popularity in state estimation tasks due to their flexibility in handling nonlinear and non-Gaussian systems~\cite{thrun2005probabilistic, doucet2001sequential}. More recent work has further expanded the applicability of particle filters to scenarios in which models of the environment are not available by \textit{learning} the components of a particle filter~\cite{jonschkowski2018differentiable, karkus2021differentiable, corenflos2021differentiable}.

An integral step in both traditional and modern particle filter algorithms is resampling, which ensures that particles with low weights are replaced by replicating particles with high weights~\cite{hol2006resampling}; resampling aims to increase the number of particles in areas intended to have high density. However, frequent resampling can lead to particle deprivation, a phenomenon where the diversity of particles diminishes over time, leaving the filter susceptible to divergence from the true state \cite{arulampalam2002tutorial, gordon1993novel}.

Recognizing these challenges, researchers have explored resampling-free particle filters to mitigate the problem of particle deprivation~\cite{pulido2019sequential, maken2022stein}. An underlying feature of these methods is the use of deterministic variational inference techniques that allow for estimation of the posterior distribution~\cite{liu2016stein}. These methods have shown potential in maintaining particle diversity over extended periods of operation. However, unlike traditional particle filters, for which convergence guarantees have been extensively studied~\cite{crisan2002survey, chopin2004central, del2006sequential}, resampling-free particle filters lack theoretical guarantees regarding their performance, especially in high-dimensional state spaces.

Our work addresses this gap, offering theoretical insights into the performance of resampling-free particle filters in high-dimensional scenarios and providing empirical evidence to support these claims. While building upon the foundations laid by previous works, we introduce key innovations that set our method apart, ensuring robust state estimation even in complex robotic systems.

\section{Designing a Resampling-free Particle Filter}
Here, we develop a resampling-free particle filter by designing a particular \textit{flow} over particles that enables particles to track the true density we wish to estimate. We then prove that our method converges to the true density even in high-dimensional state spaces. Finally, we demonstrate that our method is computationally efficient in time and memory.

\subsection{Using Flows to Track Posterior Density}
\paragraph{Setup} We consider a continuous-time hidden Markov model with hidden state $x \in \mathbb{R}^d$ and observations $y_t$, where $t$ denotes time and $d$ is the state space dimensionality. For our presentation, we consider a continuous-time setting, but our analysis and algorithm apply to discrete-time as well. Let $P_t(y|x)$ denote the probability density of observing $y$ at time $t$ such that the probability of observing a constant $y$ in an interval from $t_1$ to $t_2$ is $e^{\int_{t_1}^{t_2} \log P_t(y|x) dt}$. Denote the negative log-likelihood of observing $y_t$ at time $t$ under state $x$ as:
\begin{equation}
    L_t(x) = - \log P_t(y_t|x)
\end{equation}
We may interpret $L_t(x)$ as a loss function. We wish to find the posterior distribution $p_t(x)$ over $x$ given all observations from $y_0$ to $y_t$. By Bayes' rule, the posterior updates as:
\begin{equation}
    p_t(x) + d p_t(x) \propto p_t(x) P_t(y_t|x)^{dt}
\end{equation}
where the constant of proportionality is set such that $p_t(x) + d p_t(x)$ integrates to $1$. Intuitively, this equation states that the posterior after time $dt$ is proportional to both the prior distribution $p_t(x)$ and the likelihood of observing the data $y_t$ under state $x$ for an interval $dt$. For small $dt$, using the definition of $P_t(y|x)$, this likelihood is expressed as $e^{\int_t^{t+dt} \log P_t(y_t|x) dt} = e^{dt \log P_t(y_t|x)} = P_t(y_t|x)^{dt}$. After taking log of both sides and dividing by $dt$, this yields an expression for the time derivative of the log posterior:
\begin{equation} \label{eq:log_p_derivative}
    \frac{d}{dt} \log p_t(x) = Z_t - L_t(x)
\end{equation}
where $Z_t$ is a normalizing constant. Note that for $p_t(x)$ to always remain a probability distribution, $\frac{d}{dt} \int p_t(x) dx = 0$, which implies:
\begin{multline}
    \int \frac{d}{dt}p_t(x) dx = \int p_t(x) \frac{d}{dt} \log p_t(x) dx \\= \int p_t(x) (Z_t - L_t(x)) dx = 0
\end{multline}
Thus,
\begin{equation}
    Z_t = \mathbb{E}_{x \sim p_t} [L_t(x)]
\end{equation}
For notational convenience, we define a normalized loss $\tilde L_t(x) = L_t(x) - Z_t$ representing the loss at $x$ relative to the losses on the full density $p_t(x)$.

\paragraph{Defining a Vector Field}
Next, we design a vector field $F_t \in \mathbb{R}^d \to \mathbb{R}^d$ that when applied to posterior $p_t$, will yield a flow that tracks the posterior over time. We will first assume $F_t$ yields a flow correctly tracking $p_t$, then select a \textit{specific} $F_t$ satisfying this property. The continuity equation~\cite{lamb1924hydrodynamics} can be used to relate vector fields to the changes in density they induce:
\begin{equation}
    \dot p_t = - \nabla \cdot (p_t F_t)
\end{equation}
where $\nabla \cdot$ denotes divergence and $\dot{}$ denotes time derivative.
We may rewrite the left-hand side as:
\begin{equation}
    p_t \frac{d}{dt} \log p_t = - \nabla \cdot (p_t F_t)
\end{equation}
Substituting in Eqn~\ref{eq:log_p_derivative}:
\begin{equation}
    p_t \tilde L_t = \nabla \cdot (p_t F_t)
\end{equation}
Now, suppose there exists a time-dependent potential function $\psi_t \in \mathbb{R}^d \to \mathbb{R}$ such that its gradient with respect to $x$ equals $p_t F_t$: $\nabla \psi_t = p_t F_t$. Then:
\begin{equation}
    p_t \tilde L_t = \nabla^2 \psi_t
\end{equation}
Note that $\nabla^2$ represents the Laplacian operator. We may solve for $\psi_t$ by inverting the Laplacian as:
\begin{equation}
    \psi_t = -[p_t \tilde L_t] \star K
\end{equation}
where $\star$ denotes convolution and $K \in \mathbb{R}^P \to \mathbb{R}$ is a kernel defined as~\cite{cristoferi2017laplace}:
\begin{equation}
    K(\Delta x) = C ||\Delta x||_2^{2 - d}
\end{equation}
where $C = \frac{\Gamma(\frac{d}{2}+1)}{d(d-2) \pi^{\frac{d}{2}}}$ and $\Gamma(\cdot)$ denotes the gamma function. Then, the gradient of $\psi$ can be expressed as:
\begin{equation}
    \nabla \psi_t = -\nabla [p_t  \tilde L_t] \star K
\end{equation}
The resulting $F_t$ can be written as:
\begin{equation} \label{eq:flow}
    F_t = -\frac{\nabla [p_t  \tilde L_t] \star K}{p_t}
\end{equation}
If vector field $F_t$ is applied to density $p_t$, this yields a flow on $p_t$ that \textit{exactly} tracks the true posterior as specified by Bayes' rule (as specified by Eqn~\ref{eq:log_p_derivative}). Note that there may be other $F_t$ satisfying this property; however, this is the unique vector field such that $p_t F_t$ can be expressed as the gradient of a function $\psi_t$.

\begin{figure}
    \centering
    \includegraphics[width=0.5\textwidth]{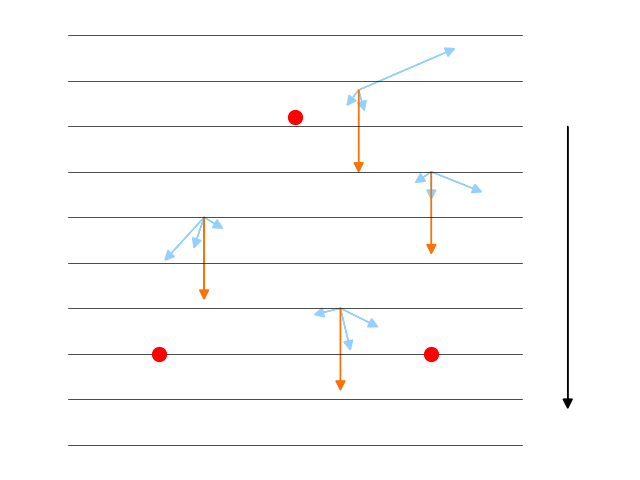}
    \caption{Example of the particle updates computed by our particle filter. Black lines indicate lines of constant likelihood $P_t$, the black arrow indicates increasing likelihood, red dots indicate particles. Sample update directions are indicated at four points: orange arrows indicate the gradient descent on the negative log-likelihood, and blue arrows indicate the attraction-repulsion force. }
    \label{fig:illustration}
\end{figure}

\paragraph{Applying the Flow to Particles}
In practice, we cannot always exactly model the posterior density $p_t$ since it often may not take an easily parametrically variable form. Instead, we use a set of $n$ particles $x^1_t, x^2_t, ... x^n_t$ to approximate the posterior density:
\begin{equation}
    p_t(x) \approx \frac{1}{n} \sum_i \delta(x - x^i_t)
\end{equation}
where $\delta(\cdot)$ denotes the $\delta$ function. From Eqn~\ref{eq:flow}, note that $p_t F_t$ may be written as:
\begin{equation}
    p_t F_t = -\nabla [p_t  \tilde L_t] \star K
\end{equation}
Plugging in the particle approximation:
\begin{equation}
    p_t(x) F_t(x) = -\nabla \left[\frac{1}{n} \sum_i \delta(x - x^i_t)  \tilde L_t(x)\right] \star K
\end{equation}
Applying the convolution on the right-hand side:
\begin{equation}
    p_t(x) F_t(x) = -\nabla \left[\frac{1}{n} \sum_i \tilde L_t(x_t^i) K(x-x_t^i)\right]
\end{equation}
Intuitively, $p_t(x) F_t(x)$ corresponds to the movement of probability density at $x$. We would like to set the movement of each particle $x_t^j$ such that its movement captures the \textit{local} value of $p_t(x) F_t(x)$ around $x^t_j$, thus capturing the correct overall movement of probability density. In other words, we would like to set $F_t(x_t^j)$ such that:
\begin{equation}
    F_t(x_t^j) = n \mathbb{E}_{\zeta}[p_t(x_t^j + \zeta) F_t(x_t^j + \zeta)]
\end{equation}
where $\zeta$ represents a random variable taking values close to $0$. Using our expression for $p_t(x) F_t(x)$:
\begin{align*}
    F_t(x_t^j) &= \mathbb{E}_{\zeta}\left[-\nabla_{x_t^j} \left[\sum_i \tilde L_t(x_t^i) K(x_t^j-x_t^i+\zeta)\right]\right]\\
    & = -\nabla_{x_t^j} \left[\sum_i \tilde L_t(x_t^i) \mathbb{E}_{\zeta}[K(x_t^j-x_t^i+\zeta)]\right]
\end{align*}
Finally, we simply approximate $\mathbb{E}_{\zeta}[K(x_t^j-x_t^i+\zeta)]$ as:
\begin{equation}
    \mathbb{E}_{\zeta}[K(x_t^j-x_t^i+\zeta)] \approx C (||x_t^j-x_t^i||_2^2 + \gamma^2)^{\frac{2-d}{2}}
\end{equation}
where $\gamma$ is a small constant added to approximate the fact that $x_t^j-x_t^i+\zeta$ will always be non-zero with probability $1$ even when $x_t^j$ and $x_t^i$ are equal. This yields the following final flow for particle $x_t^j$:
\begin{align} \label{eqn:final}
    F_t(x_t^j) &= -C \nabla_{x_t^j} \left[\sum_i \tilde L_t(x_t^i) (||x_t^j-x_t^i||_2^2 + \gamma^2)^{\frac{2-d}{2}} \right] \nonumber\\
    &= -C \gamma^{2-d} \nabla_{x_t^j} \tilde L_t(x_t^i) \nonumber \\&- C \sum_{i}  \frac{ (d-2) \tilde L_t(x_t^i)}{(||x_t^j-x_t^i||^2_2 + \gamma^2)^{\frac{d}{2}}}  (x_t^i-x_t^j)
\end{align} 
Intuitively, the first term of the update simply performs gradient descent on the loss $\tilde L_t$. The second term corresponds to an attraction-repulsion force that shifts point $x_t^j$ \textit{towards} point $x_t^i$ if the loss on point $x_t^i$ is favorable (i.e. $<0$) and away from point $x_t^i$ if the loss is unfavorable (i.e. $>0$). Please see Fig~\ref{fig:illustration} for an illustration of our update rule; note that the attraction-repulsion force may result in an update \textit{opposite} to the direction of gradient descent.

\subsection{Theoretical Guarantees of Convergence to Posterior}

The particle ensemble approximation of the posterior introduces an error with the true posterior. In this section, we quantify this approximation error theoretically.

Let us denote the true posterior by $p_t$ and our particle ensemble approximation by $q_t$. We assume $p_t$ flows according to the true vector field $F_t$ while $q_t$ flows under an approximated vector field $\tilde F_t$. For simplicity, we assume that both $p_t$ and $q_t$ are discrete and consist of $N$ points each, denoted $\{x_t^i\}_{i=1}^N$ for $p$ and $\{z_t^i\}_{i=1}^N$ for $q$, though the bound can be generalized to the continuous case by taking the limit in $N$. 

We use the Wasserstein distance as our distance measure. The Wasserstein distance between $p_t$ and $q_t$ is defined as
\begin{equation}
W(p_t, q_t) = \min_\pi \sum_{i=1}^N d(x^i_t, z^{\pi(i)}_t)
\end{equation}
where the expression is defined over a metric space with distance measure $d$, and the minimum is taken over all permutations $\pi$ on $[N]$. Let $\pi_t$ denote the minimizing $\pi$ for the distributions at time $t$. Additionally, suppose that $d$ has the shift-invariant property; that is,
\begin{equation}
d(x+\delta, z+\delta) = d(x, z)
\end{equation}
for all $x, z, \delta$. 

We present the following theorem on the approximation error.
    
\begin{theorem}
    Let $L_F$ and $L_d$ be the Lipschitz constants of $F$ and $d$, respectively. Further, suppose that the maximum pointwise discrepancy between $F$ and $\tilde F$ is bounded by $\frac{\varepsilon}{N}$. Then, 
    \begin{equation}
    W(p_t, q_t) \leq \left(W(p_0, q_0) + \frac{\varepsilon}{L_F}\right) e^{L_d L_f t} - \frac{\varepsilon}{L_F}.
    \end{equation}
\end{theorem}
\begin{figure*}  
  \begin{subfigure}{0.5\textwidth} 
    \includegraphics[width=\linewidth]{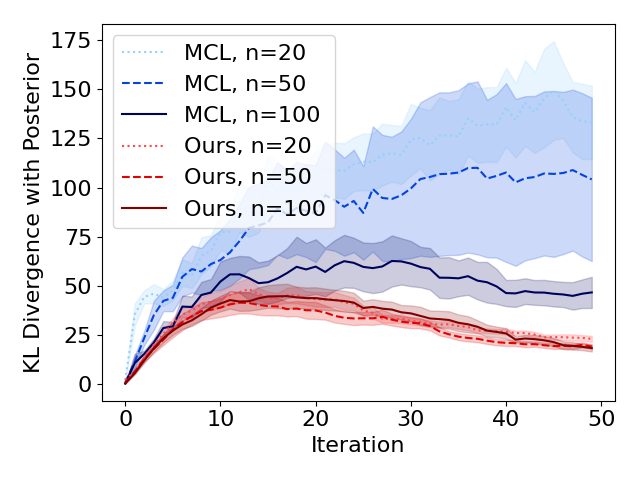}
  \end{subfigure}
  \begin{subfigure}{0.5\textwidth} 
    \includegraphics[width=\linewidth]{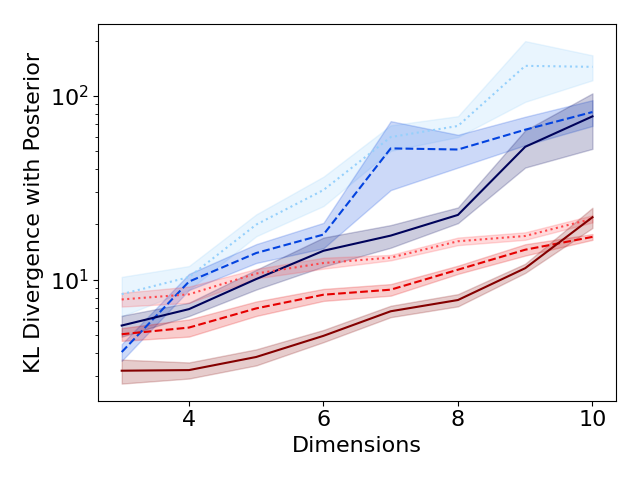}
  \end{subfigure}
  \caption{Comparison of the performance (measured as KL divergence with the true posterior distribution) of our particle filter with standard Monte Carlo Localization (MCL) on a synthetic localization problem with adjustable dimensionality. $n$ denotes the number of particles. Margins indicate standard errors over $10$ trials. [Left] Performance over the course of particle filter iterations on a $10$-dimensional problem. [Right] Performance at the final iteration over varying dimensions.}
  \label{fig:toy_problem}
\end{figure*}
\begin{proof}
Assume that $p_t$ follows the true flow $F_t$ and $q_t$ follows the approximate flow $\tilde F_t$. We then  bound the Wasserstein distance at time $t+dt$ using the Wasserstein distance at time $t$.
\begin{align*}
W(p_{t+dt}&, q_{t+dt}) \\
&\leq \sum_i d(x^i_t +  F_t(x^i_t) dt, z^{\pi_t(i)}_t + \tilde F(z^{\pi_t(i)}_t) dt)\\
&= \sum_i d(x^i_t +  [F_t(x^i_t) - \tilde F_t(z^{\pi_t(i)}_t)] dt, z_t^{\pi_t(i)})\\
&\leq \sum_i d(x_t^i, z_t^{\pi_t(i)}) + L_d ||F_t(x^i_t) - \tilde F_t(z_t^{\pi_t(i)})|| dt
\end{align*}
where we applied the shift-invariant property of $d$. We then rewrite the above as 
\begin{align*}
    \sum_i &d(x_t^i, z_t^{\pi_t(i)})\\
    &+L_d ||F_t(x_t^i) - F_t(z_t^{\pi_t(i)}) + F_t(z_t^{\pi_t(i)}) - \tilde F_t(z_t^{\pi_t(i)})|| dt
\end{align*}
which can be bounded by 
\begin{equation}
 \sum_i d(x^i_t, z_t^{\pi_t(i)}) + L_d (L_F d(x_t^i, z_t^{\pi_t(i)}) + \varepsilon) dt
\end{equation}
and rewritten as 
\begin{equation}
(1 + L_d L_F dt) W(p_t, q_t) + L_d \varepsilon dt.
\end{equation}
We thus obtain a differential equation as our final inequality:
\begin{equation}
W(p_{t+dt}, q_{t+dt}) \leq (1 + L_d L_F dt) W(p_t, q_t) + L_d \varepsilon dt
\end{equation}
which solves via Grönwall's inequality~\cite{gronwall} to
\begin{equation}
W(p_t, q_t) \leq \left(W(p_0, q_0) + \frac{\varepsilon}{L_F}\right) e^{L_d L_f t} - \frac{\varepsilon}{L_F}.
\end{equation}
\end{proof}
Our result reveals that the error at time $t$ scales linearly with the error at time $0$ (as $n$ or $d$ is varied). Critically, this means that even in very high-dimensional state spaces, if a strong initial approximation of the prior distribution $p_0$ is provided, then the approximated posterior $q_t$ is guaranteed to track $p_t$ up to some constant error term and scale factor. In other words, if $W(p_0, q_0)$ is bounded by a dimension-independent constant $K$, then $W(p_t, q_t)$ can also be bounded by a dimension-independent constant: \textit{our particle filter avoids the curse of dimensionality}. We note that $W(p_0, q_0)$ may generally scale with the dimensionality of the state space: in higher dimensional spaces, more particles may be necessary to approximate any distribution to the same precision. However, this dimensionality dependence is purely due to the difficulty of approximating distributions in high-dimensional spaces, and is unavoidable for \text{any} particle filter. Our particle filter adds \textit{no further error} in high-dimensional spaces relative to low-dimensional ones.

Moreover, unlike classical convergence results on standard (resampling-based) particle filters~\cite{crisan2002survey}, due to the deterministic nature of our particle filter (it avoids any sampling), our convergence results are not probabilistic. Instead, convergence is always guaranteed to hold.

Interestingly, note that we may directly apply the classical convergence results~\cite{crisan2002survey} at time step $0$ of our algorithm, and achieve the same convergence rate (up to constant terms and scale factors) at future timesteps $t$. For instance, the classical $O(\frac{1}{n})$ convergence rate of $(\mathbb{E}_p[f(x)] - \mathbb{E}_q[f(x)])^2$ for any bounded function $f$ directly applies to our filter as well, including the dimension-independence of the convergence rate.

\subsection{Computational Efficiency}
We present our full algorithm in Algorithm~\ref{algo:resampling-free-particle-filter}. Observe that each step of the algorithm requires looping over all $n$ particles to update each of them, and each particle's update step requires computing displacements with all other particles. Thus, our algorithm's runtime is $O(n^2T)$ where $T$ is the number of total timesteps and $n$ is the number of particles. Note that the quadratic dependence on $n$ can be computationally expensive for large numbers of particles. However, as we have previously shown, if the initial error of our particle filter is bounded, then the error in the posterior is \textit{independent of dimension}; thus, the required number of particles to achieve a particular error rate also does not scale with dimension.

Storing each particle requires $O(d)$ memory, resulting in a memory consumption of $O(nd)$, which is the minimum that can be reasonably expected to represent the particles.

\begin{algorithm}
\caption{Resampling-free Particle Filter}
\label{algo:resampling-free-particle-filter}
\begin{algorithmic}[1]
\Require Initial particles $\{x^i_0\}_{i=1}^n$, time step $\Delta t$, small constant $\gamma$, total timesteps $T$, negative log-likelihood function $L_t$, constant $C$, dimensionality $d$
\For{$t = 1, 2, \ldots, T$}
    \State Compute $L_t(x^i_t)$ for all $i$
    \State $Z_t = \frac{1}{n} \sum_i L_t(x^i_t)$
    \For{$j = 1, 2, \ldots, n$}
        \State $\tilde{L}_t(x^j_t) = L_t(x^j_t) - Z_t$
        \State $x_{t+1}^j = x_{t}^j  -C \gamma^{2-d} \nabla_{x_t^j} \tilde L_t(x_t^j)$
        \State $x_{t+1}^j = x_{t+1}^j - C \sum_{i}  \frac{ (d-2) \tilde L_t(x_t^i)}{(||x_t^j-x_t^i||^2_2 + \gamma^2)^{\frac{d}{2}}}  (x_t^i-x_t^j)$
    \EndFor
\EndFor
\end{algorithmic}
\end{algorithm}

\begin{figure*}  
  \begin{subfigure}{0.5\textwidth} 
    \includegraphics[width=\linewidth]{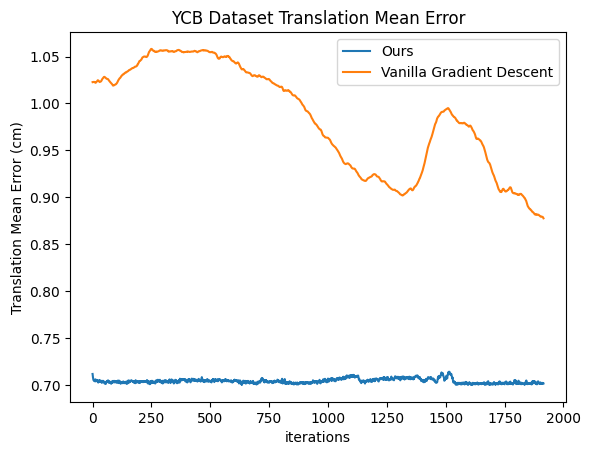}
  \end{subfigure}
  \begin{subfigure}{0.5\textwidth} 
    \includegraphics[width=\linewidth]{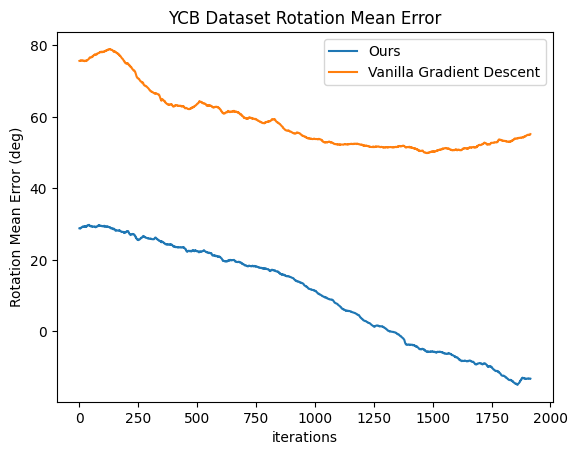}
  \end{subfigure}
      \caption{Performance Analysis of 6D Pose Estimation Methods on the "Cracker Box" Model: This figure presents a comparative analysis of the translation and rotation mean error between our proposed method and the conventional gradient descent approach. The evaluation is based on 1912 RGB video sequences of the "cracker box" model. Both methods were tested under identical configurations, using 80 particles. The mean error is computed over the \(x, y, z\) coordinates and averaged for both translation and rotation. Specifically, the translation error measures the deviation in centimeters from the ground truth translation, while the (signed) rotation error quantifies the angular difference in degrees from the true rotation. Note that the rotation error can be positive or negative, with the optimal value being $0$.}
  \label{fig:ycb_problem}
\end{figure*}

\section{Experiments}
To assess the effectiveness of our approach on high-dimensional localization, we conduct experiments on a synthetic high-dimensional localization task with a variable state-space dimensionality and $6D$ pose estimation.

\subsection{Synthetic Localization}
\paragraph{Setting}
We consider the following log-likelihood function:
\begin{equation}
    L_t(x) = \frac{1}{2} ((x - u)^T \xi_t)^2
\end{equation}
where $u \in \mathbb{R}^d$ is an \textit{unknown} parameter drawn from $N(0, I)$ and $\xi_t \in \mathbb{R}^d$ are drawn iid from $N(0, I)$. Due to the simple form of the likelihood, we may analytically solve for the true posterior over $x$. Specifically, assuming a prior of $N(0, I)$ over $x$, the corresponding expected log posterior after $t$ timesteps may be expressed as:
\begin{equation}
    \log p_t(x) = -\frac{1}{2} ||x||_2^2 - \frac{t}{2} ||x - u||_2^2
\end{equation}
We compare our particle filter and a baseline of standard Monte Carlo Localization (MCL)~\cite{fox1999monte}. For MCL, we add Gaussian noise from $N(0, \epsilon I)$ as the motion model: $P(\xi_{t+1}-\xi_t|\xi_t) \sim N(0, \epsilon I)$, where $\epsilon$ is a hyperparameter. For both methods, we conduct grid searches over each method's relevant hyperparameter over $5$ orders of magnitude and report results for the best setting. We run the particle filter for $50$ iterations and evaluate performance at each iteration $t$ using KL divergence between a multivariate Gaussian fit to the particles and the true posterior at time $t$. We also assess performance as a function of the state space dimensionality $d$.

\paragraph{Results}
In Fig~\ref{fig:toy_problem}, we find that for both particle filters, the KL divergence initially increases as the true posterior shifts away from the prior distribution, but then decreases as the true posterior concentrates its density. Our method outperforms MCL over all iterations tested. Strikingly, while MCL worsens greatly as the number of particles decreases, our method maintains its performance. Moreover, our method much more gracefully handles higher dimensional state spaces, achieving roughly a $\times 10$ reduction in KL divergence compared to MCL at the highest dimension tested.

\subsection{6D Pose Estimation}

\paragraph{Setting}
6D pose estimation, which involves estimating the 3D rotation and translation of objects relative to the camera, is central to many robotic tasks such as manipulation and navigation, enabling efficient grasp planning, obstacle avoidance, and more. Historically, local-feature or template-matching techniques have been the backbone for estimating the 6D pose of objects. These methods typically involve matching image features against pre-generated templates or features of a 3D object model, thereby recovering the object's pose~\cite{deng2019pose}. 

We evaluate the performance of our particle filter to perform pose estimation. Here, note that each particle corresponds to a pose estimate for the object in question. We adapt the approach in \cite{deng2019pose} to factorize the posterior into two components: the 3D translation and the 3D rotation of the object; each particle corresponds to a specific translation and rotation. Thus, each particle has 6 dimensions represented as a combination of translation and rotation. 

We use the same approach as in~\cite{deng2019pose} to compute the likelihood of a particular observation $y$ being consistent with a pose $(T, R)$ consisting of translation $T$ and rotation $R$. Critically, we assume access to a \textit{reference} object for which we have observations under a variety of rotations. We use a pre-trained autoencoder that allows us to compare the similarity of two objects: the observation $y$ under the candidate translation $T$ and the appearance of reference object under candidate rotation $R$.  If the pose is more likely, then the reference object at rotation $R$ will be similar to the observation translated with $T$. Thus, we set generate a likelihood function as follows:
\begin{equation}
    P(y|(T, R)) \propto S(g(crop(y, T)), g(\bar y^R))
\end{equation}
where $g$ is a pre-trained encoder mapping images to a vector code, $crop(y, T)$ applies a crop to image $y$ to isolate an object in the image assuming it is located at translation $T$, $\bar y^R$ denotes a reference image of the object under rotation $R$, and $S$ is a similarity function. Given this likelihood function, we run our particle filter as described in Algorithm~\ref{algo:resampling-free-particle-filter} using randomly initialized initial pose estimates. We compare our particle filter with a baseline of simply applying gradient descent on the negative log-likelihood function (corresponding to ignoring the attraction-repulsion force in our update equation Eqn~\ref{eqn:final}).

\paragraph{YCB dataset results}
Our experiments were conducted on the YCB Video dataset~\cite{xiang2017posecnn}, specifically on RGB video sequences of household objects, both textured and textureless. Each video sequence consists of a series of frames in which the object may have varying pose. The objects in this dataset are annotated with their 6D poses at each frame. For our experiment, we used the "cracker box" object which consisted of $1912$ RGB video frames. We evaluated the performance of our particle filter at estimating the object's pose over time using $80$ particles. Specifically, we measured the error between the true pose of the object and mean pose estimated by our particles.

As seen in Figure~\ref{fig:ycb_problem}, our results revealed a marked enhancement in 6D pose estimation accuracy using PoseRBPF as opposed to gradient descent. Our proposed method consistently yielded a diminished error when compared to the gradient descent baseline. Both in terms of rotation and translation error rates, our algorithm exhibited superior efficiency in terms of number of iterations, arriving at values closer to the ground truth faster than gradient descent. This analysis underscores the potential of our approach in handling pose estimation tasks.

\section{Conclusion}
We develop a novel resampling-free particle filter designed to circumvent particle deprivation issues that arise when applying traditional particle filters in high-dimensional state spaces. Through theory and experiments, we demonstrate that our approach avoids the curse of dimensionality and is practically effective in high-dimensional localization. We believe particle filters explicitly designed for high-dimensional state spaces will be critical in practical applications as robots are increasingly designed with more degrees of freedom.

\addtolength{\textheight}{-12cm}   

\bibliographystyle{ieeetr}
\bibliography{ref}

\end{document}